\documentclass{article}
\usepackage{spconf,amsmath,graphicx}
\usepackage{booktabs}
\usepackage{times}
\usepackage{epsfig}
\usepackage{graphicx}
\usepackage{amsmath}
\usepackage{amssymb}
\usepackage{enumitem}
\usepackage{tablefootnote}
\usepackage{multirow}
\usepackage{capt-of}
\usepackage{subcaption}

\usepackage[table]{xcolor}% http://ctan.org/pkg/xcolor

\title{Thermal to Visible Image Synthesis under Atmospheric Turbulence}

\name{Kangfu Mei \hspace{0.1cm}, Yiqun Mei\hspace{0.16cm} and \hspace{0.16cm} Vishal M. Patel \thanks{This work was supported by NSF CAREER award 2045489.}}
\address{Dept. of Electrical and Computer Engineering, Johns Hopkins University, MD, USA \\\texttt{\{kmei1, ymei7, vpatel36\}@jhu.edu}}
\begin{document}
 \ninept
\maketitle
\begin{abstract}
In many practical applications of long-range imaging such as biometrics and surveillance, thermal imagining modalities are often used to capture images in  low-light and nighttime conditions.  However, such imaging systems often suffer from atmospheric turbulence, which introduces severe blur and deformation artifacts to the captured images.  Such an issue is unavoidable in long-range imaging and significantly decreases the face verification accuracy.   In this paper, we first investigate the problem with a turbulence simulation method on real-world thermal images. An end-to-end reconstruction method is then proposed which can directly transform thermal images into visible-spectrum images by utilizing natural image priors based on a pre-trained StyleGAN2 network.  Compared with the existing two-steps methods of consecutive turbulence mitigation and thermal to visible image translation, our method is demonstrated to be effective in terms of both the visual quality of the reconstructed results and face verification accuracy. Moreover, to the best of our knowledge, this is the first work that studies the problem of thermal to visible image translation under atmospheric turbulence.

%Thermal images that capture body heat emission information have shown promising priority in face verification applications compared with images captured in visible-spectrum, especially at low-light and nighttime conditions. However, thermal imaging can be easily affected by atmospheric turbulence, which introduces severe blur and deformation artifacts to the original images. Such an issue is unavoidable at the long-range imaging task and significantly decreases the face verification accuracy. In this paper, we first investigate the problem with reasonable turbulence simulation technology on real-world thermal images. An end-to-end reconstruction method is then proposed to learn to transform thermal images into visible-spectrum images directly without intermediate steps, and it enables a pre-trained StyleGAN2 to utilize its natural image priors. Compared with the existing two-steps methods of consecutive turbulence mitigation and thermal-visible images translation, our method is demonstrated to be effective in terms of both the visual quality of reconstructed results and face verification accuracy. Moreover, to the best of our knowledge, this is the first work that discusses the thermal-visible image translation problem under turbulence, which may serve as a strong baseline for future works focused on the similar invisible-spectrum translation task affected by turbulence.
\end{abstract}
\section{Introduction}
In many applications of long-range imaging, such as surveillance, we are faced with a scenario where we have to determine the identity of a person appearing in the captured imagery degraded by atmospheric turbulence.
One way to deal with this problem is to develop methods that can remove the effect of turbulence from images.
However, restoring images degraded by atmospheric turbulence is difficult since it causes images to be geometrically distorted and blurry.
Such a scenario becomes more challenging when the images are captured from thermal modality, for example, under low-light and nighttime conditions.
In this work, we investigate the task where we have to synthesize visible images from the captured thermal images which are degraded by atmospheric turbulence.

\begin{figure}[htbp]
    \centering
    \includegraphics[width=.9\linewidth]{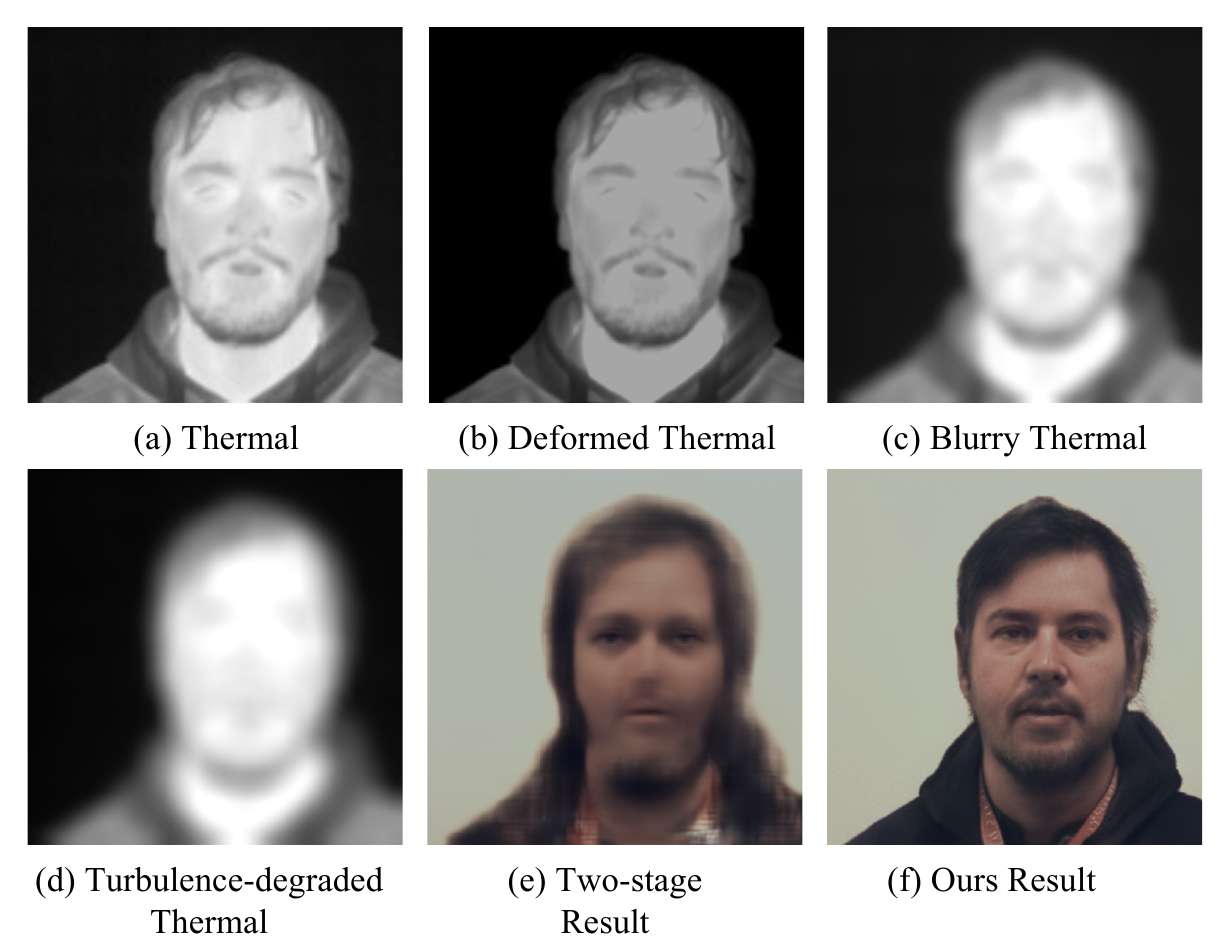}
   \vskip-5pt
   \caption{Visualization of the degraded thermal image under atmospheric turbulence and its corresponding reference image. Most high-frequency details are missing in the thermal image under turbulence compared with the deformed only and the blur only images.}
   \label{fig:teaser}
\end{figure}

Existing efforts on thermal-to-visible recognition algorithms can be divided into two main categories: recognition via \textit{feature extraction}~\cite{klare2012heterogeneous,riggan2016optimal,gong2017heterogeneous} and \textit{recognition via synthesis}. In the first category, methods aim to find a common feature space containing discriminative information, where images across different modality but belonging to the same person can be mapped closer under certain relevance measurements. Classical approaches includes kernel prototype similarities~\cite{klare2012heterogeneous}, partial least squares~\cite{choi2012thermal} and handcrafted feature descriptors such as SIFT and HOG \cite{riggan2016optimal}. Recently deep learning-based methods have shown to be effective in solving various computer vision tasks,  motivating more recent methods to explore it for superior performance. Common techniques include deep metric learning~\cite{saxena2016heterogeneous} to reduce the domain gap between thermal and visible modality or leveraging convolutional neural networks (CNNs) to extract domain-invariant features~\cite{fondje2020cross,he2017learning,he2018wasserstein,iranmanesh2018deep}.

Recently, \textit{recognition by synthesis}~\cite{mallat2019cross,zhang2019synthesis,di2018polarimetric,di2019polarimetric,immidisetti2021simultaneous} has been used to address the problem of heterogeneous face recognition since any off-the-shelf face recognition method can be seamlessly applied on the translated visible images.  Riggan et al.~\cite{riggan2018thermal} synthesized images by leveraging both global and local regions, resulting in better discriminative quality. Later, Mallat et al.~\cite{mallat2019cross} introduced cascaded networks to gradually refine the generated images. Several recent methods leverage Generative Adversarial Networks (GANs) to further improve the perceptual quality of the synthesized images. Specifically, Zhang et al.~\cite{zhang2019synthesis} proposed GAN-VFS that learns to jointly optimize visible feature estimation and facial reconstruction. Di et al.~\cite{di2019polarimetric} and Immidisetti et al.~\cite{immidisetti2021simultaneous} adopt mulltiple self-attention modules into their GANs to allow long-range correlation modeling, which further enhance the synthesis quality. Our work falls into the later category, which learns to synthesis visible images from thermal modality.

\begin{figure*}[htbp]
    \centering
    \includegraphics[width=.85\linewidth]{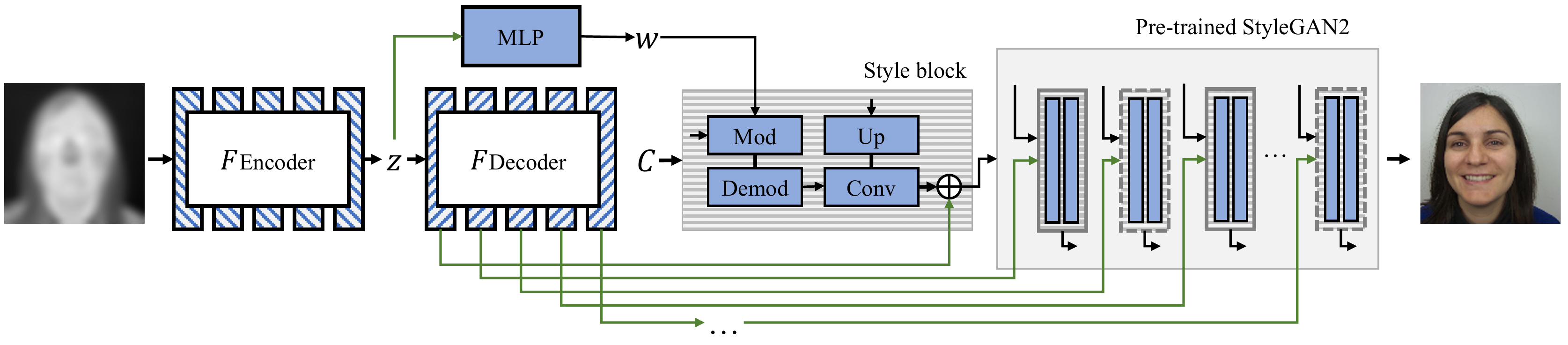}
    \vskip -10pt\caption{An overview of the proposed end-to-end network for thermal to visible image synthesis  under atmospheric turbulence. The network ensembles a pre-trained StyleGAN2 and learns to project the thermal images into the style-space of a pre-trained StyleGAN2. Resulting from the rich generative priors of StyleGAN2, the network can produce sharper visible images in an end-to-end manner without processing turbulence or thermal images separately.}
    \label{fig:pipeline}
\end{figure*}

The visual quality of imaging through turbulence suffers distortion from both the blur and deformation operations in the pixel space.
The physical model corresponding to turbulence degradation has been established in~\cite{kolmogorov1991local, tatarski2016wave, fried1965statistics, fried1966optical, fried1978probability}.  It has been reformulated and simplified in the turbulence mitigation works~\cite{yasarla_learning_2020, lau_atfacegan_2021, yasarla_learning_2021, nair_confidence_2021} as 
\begin{equation}
 y = T(H(x)) + \xi,
\end{equation}
where ${y}$ is the observed turbulence degraded image, $x$ is the clean image, $\xi$ is the additive noise, $T$ is the turbulence degradation operator, and $H$ corresponds to visible to thermal operator.
Practical ways for simulating turbulence-like images include physics-based parameter simulation, e.g., Chimitt et al.~\cite{chimitt2020simulating} and Mao et al.~\cite{mao2021accelerating}, and visual effects simulation, e.g., Lau et al.~\cite{lau_atfacegan_2021} and Yasarla et al.~\cite{yasarla_learning_2021}.
Though large differences exist in these two ways of simulating turbulence, their results are similar in the sharpness and color bias.   Hence, most mitigation methods make use of them without any specific configurations.  In this work, we simulate turbulence degraded thermal images according to Mei and Patel~\cite{mei2021ltt}, which combines multiple random blur and noise with Elastic deformation augmentation.
We empirically find that the parameters of simulation following the 300 meters long-range distance configuration leads to the best simulation effects in the thermal images.

Turbulence mitigation is another emerging topic beyond simulation.
One straightforward way is build upon two-step mitigation and has been widely applied in recent deep-learning based methods, e.g., TDRN~\cite{yasarla_learning_2021}, ATFaceGAN~\cite{lau_atfacegan_2021}.
Such methods tend to process the turbulence degraded image with the deformation correction module and deblurring module, and the two results are then fused to obtain the final result.
However, recent research~\cite{mei2021ltt} shows that learning mitigation in an end-to-end fashion avoids error propagation which often happens in the two-step process.  Inspired by \cite{mei2021ltt}, in this work, we propose a novel network for thermal to visible image reconstruction under turbulence.
The network ensembles a pre-trained GAN and utilizes the GAN prior for learning the reconstruction.
Benefiting from the generative priors, we find that such a network is able to simultaneously restore and translate a turbulence degraded thermal image into a high-quality visible image.

\section{Proposed Method}
The observation model we follow is $\tilde{I}=T(H(I)) + \xi$ that is mentioned before.  Given a thermal image $\tilde{I}$ captured under atmospheric turbulence, we propose to learn to reconstruct a visible image $I$ using a deep neural network $G_\theta (\cdot)$ and optimize its parameters $\theta$ according to an objective with the ground truth $I$.   The goal of the network  $G$ is to simultaneously restore and translate a thermal image into a visible image.  Specifically, the objective includes the adversarial loss of GAN which is combined with the pixel-wise loss, perceptual loss in the pre-trained VGG19 network $\phi(\cdot)$, and identity preserving loss defined in the pre-trained face recognition network $\eta(\cdot)$ as
\begin{equation}
\begin{split}
  \mathcal{L} (G) & = - \lambda_{adv} \mathbb{E}_{G(\tilde{I})} \operatorname{softplus}(D(G(\tilde{I}))) \\
  & +  \| I - G(\tilde{I}) |\ + \lambda_{per} \|\phi(I) - \phi(G(\tilde{I})) \| \\
  & + \lambda_{id} \|\eta(I) - \eta(G(\tilde{I})) \|,
\end{split}
\end{equation}
where the $\lambda_{adv}$, $\lambda_{per}$ and $\lambda_{id}$ are the weights of the adversarial loss, perceptual loss, and identity preserving loss, respectively.
The overall architecture in illustrated in Figure~\ref{fig:pipeline}.

The proposed project module is build upon an encoder-decoder network which takes thermal images $\tilde{I}$ as input and outputs a latent code $z$ and a set of modulation features.  In what follows, we provide details of our network.

%The details regarding thermal images under turbulence is first introduced in Section \ref{sec:encoder}.  Then Section~\ref{} presents an overview of the architecture for reconstructing the clear images given the latent code and the modulation features.

\subsection{Thermal-Turbulence Projection Module}
\label{sec:encoder}
To find the latent code $z$ and modulation features that correspond to the clear images, the projection module should be capable enough to capture both the identity information and local structures. 
However, extracting local information from thermal images is pretty challenging since most of the details are distorted.
We propose to utilize a multi-scale encoder-decoder network for extracting features from $\tilde{I}$, where the number of encoder and decoder layers and their resolutions of the output follows the configuration of StyleGAN2 \cite{karras2020analyzing}.
Denoting the first feature extraction layer of the encoder part as $\mathbf{E}_0(*)$, we have the shallow feature $F_0$ as $F_0 = \mathbf{E}_0(\tilde{I}).$
%\begin{equation}
   % F_0 = \mathbf{E}_0(\tilde{I}).
%\end{equation}
In particular, $n$ number of encoder layers with a pooling operation of scale 1/2 are used to extract multi-scale features and preserve the details as follows
\begin{equation}
    F_i = \mathbf{E}_i(\operatorname{Pooling}(F_{i-1})), i\in \{1,2,\dots, n\}.
\end{equation}
The final output of $F_n$ is then taken as the predicted latent code $z$ for projection. 
In order to preserve the details of the reconstructed images at different scales, the extracted features $\{F_1, F_2, \dots, F_n\}$ are then processed by $n$ decoder layers $\mathbf{D}_i(\cdot)$ as
\begin{equation}
    \bar F_i = \mathbf{D}_i(\operatorname{Deconvolution}(F_{i-1})) + F_i, i\in \{1,2,\dots, n\}.
\end{equation}
These multi-scale features $\{\bar F_1, \bar F_2, \dots, \bar F_n \}$ are then applied as the feature modulation parameters for gradually correcting the style features of a pre-trained StyleGAN2 at its generation process.
Based on the aforementioned two types of encoder and decoder layers, the proposed projection module can learn to project the thermal images under turbulence into the natural image space encoded by StyleGAN2.
Here we use green lines in Figure~\ref{fig:pipeline} to denote the connections between the projection module and the pre-trained StyleGAN2 for clarification.
Note that at the learning process, only the parameters of the projection module are updated according to the gradients, while the parameters of the pre-trained StyleGAN2 are fixed.

\subsection{Image Reconstruction Module}
As mentioned in Section~\ref{sec:encoder}, the parameters of the pre-trained StyleGAN2 are fixed during the entire learning procedure, and thus its output always fits the natural image statistics given an arbitrary latent code. 
Such property significantly simplifies the reconstruction learning since the latent space is limited to the manifold corresponding to natural images only.
However, since the generation process of StyleGAN2, i.e., mapping random latent code to natural images is stochastic, ensuring the identity consistency of reconstructed images can be difficult.
To overcome this issue, we leverage multi-scale features $\{\bar F_1, \bar F_2, \dots, \bar F_n\}$ produced by the decoder to modulate the features of StyleGAN2 at generation.
In particular, for each output layers $\mathbf{L}_i(\cdot)$ of StyleGAN2 at each resolution, the original procedure takes features $\hat F_{i-1}$ extracted from noise and generates the features in the next level as $\hat{F}_{i} = \mathbf{L}_i(\hat F_{i-1}, \epsilon),$
%\begin{equation}
   % \hat{F}_{i} = \mathbf{L}_i(\hat F_{i-1}, \epsilon),
%\end{equation}
where $\epsilon$ is the noise corresponding to normal distribution.
In contrast, our modified version modulates the generation process as
\begin{equation}
    \begin{split}
        \bar F_i^{\mathrm{mean}}, \bar F_i^{\mathrm{std}} &= \operatorname{Split}(\bar F_i), \\
        \hat{F}_{i} &= (\mathbf{L}_i(\hat F_{i-1}, \epsilon) + \bar F_i^{\mathrm{mean}}) * \bar F_i^{\mathrm{std}},
    \end{split}
\end{equation}
where $\operatorname{Split}(\cdot)$ divides the decoded feature $\bar F_i$ into two modulate parameters $\bar F_i^{\mathrm{mean}}$ and $\bar F_i^{\mathrm{std}}$ at the channel dimension.
We empirically find that such modulation  is able to correct the features during generation, and it helps to preserve details at reconstruction.  Following such a modulation process, the final output of the pre-trained StyleGAN2 can preserve both the identity-related details and the natural image statistics.

\section{Experiments}
In this section, we conduct experiments to evaluate our approach against existing state-of-the-art approaches. We select two commonly used thermal-visible datasets: the VIS-TH~\cite{mallat2018benchmark} dataset and the ARL-VTF~\cite{poster2021large} dataset. In the following, we first briefly introduce them. Then we describe the evaluation metrics, and training and implementation details. Finally, we present both quantitative and qualitative results to showcase the superiority of our method.

\subsection{Dataset and Evaluation Metrics}
%So far, there are no standardized protocols in this field. Existing methods report results trained and tested on custom datasets/splits. This paper selects two common datasets (VIS-TH~\cite{mallat2018benchmark} and ARL-VTF~\cite{poster2021large}) where high-resolution VIS images are available. To study the effect of data restriction, we intentionally choose one dataset to be more challenging than the other. The precise dataset splits will be released. We create VIS-TH image pairs by cropping $512\times512$ and $128\times128$ face regions respectively. 

%\noindent\textbf{VIS-TH} is a challenging dataset containing data from 50 subjects.  Images from each subject contain 21 faces varying significantly in pose, expression and light conditions. VIS-TH images are captured via a dual-sensor camera in LWIR modality and thus naively aligned. We construct the training set by randomly selecting data from 40 subjects. The remaining data from 10 subjects are used as the testing set. 
\noindent\textbf{VIS-TH} is a challenging Visible-Thermal dataset which is captured in the Long Wave Infrared  (LWIR) modality. It contains data from 50 subjects.  Images from each subject contain variations in expression, pose and illumination conditions.  The paired thermal and visible images are captured by a dual-sensor camera and thus are well-aligned. We randomly select data from 35 subjects for training, data from 5 subjects for validation. The remaining 10 subjects are used for testing.

\noindent\textbf{ARL-VTF} is a popular dataset for thermal-to-visible face verification, consisting of data from 220 identities. Images for each subject vary only in expressions. Annotations are provided for alignment. We construct the training set by randomly selecting 160 subjects. We also randomly selects 40 subjects for evaluation and use the rest 20 subjects for testing. This results in 3,200 training pairs, 400 validation pairs and 985 testing pairs. We apply a simple color adjustment to mitigate overexposure over the VIS modality. 

%provides data in LWIR modality from 220 subjects with annotations for alignment. We create a  dataset by randomly selecting a subset of 160 subjects with variations only in expressions as the training set, 20 subjects' data as the validation set, and 40 subjects' data as the testing set. The resulting data split contains 3,200 training pairs, 400 validation pairs, and 985 testing pairs. The color adjustment is applied to mitigate overexposure of the VIS images.  

%The presented  also show strong overexposure, a property unsuitable for image quality assessment. We solve this by applying a simple 

\subsection{Evaluation Metrics.} To best demonstrate the effectiveness of our approach, we report results with both face verification metrics and image quality measurements. Following~\cite{duan2020cross}, we report Rank-1 accuracy, Verification Rate (VR) @ False Accept Rate (FAR)=$1\%$ and VR@FAR=$0.1\%$ for evaluating face recognition. We create the gallery set by selecting one visible image for each subject and use all thermal images as the probe set. 
For image quality, perceptual metrics LPIPS~\cite{zhang2018unreasonable}, NIQE~\cite{mittal2012making}, identity metric Deg (cosine distance between LightCNN~\cite{wu2018light} features), and pixel-wise PSNR and SSIM~\cite{wang2004image} are reported for comparison.

\subsection{Implementation and Training Details}
For reconstructing faces, we leverage the pre-trained StyleGAN2~\cite{karras2020analyzing}. The projection module for encoding styles and modulation features contains 7 downsample layers and 7 upsample layers. At the lowest level, features have a spatial dimension of $4\times 4$. The size of all convolution filters is set to $3\times 3$. During training, each input batch contains 4 thermal images. We set $\lambda_{adv}=1$; $\lambda_{per}=\lambda_{id}=10$. To optimize the parameters, we adopt the Adam~\cite{kingma2015adam} optimizer with $\beta_{1}=0.9$, $\beta_{2}=0.999$ and $\epsilon=1e-8$. The initial learning rate is set equal to 2e-3 and reduces to a half after 140K iterations. The training stops at 150K iterations. We implement the proposed model using PyTorch on Nvidia RTX8000 GPUs.

\subsection{Turbulence Data Simulation}
The simulation method is inspired by Mei and Patel~\cite{mei2021ltt}, which is originally proposed for turbulence simulation on visible-spectrum images.
We experimentally find that such a simulation method is also suitable for thermal images.
Specifically, we applied the random blur and elastic deformation on the thermal images, where \emph{isotropic} and \emph{anisotropic} Gaussian kernels are used with a fixed blur kernel size $11$ and sampled blur $\sigma$ from $[1, 11]$.
For the elastic deformation, we empirically choose the parameters $\alpha$ and $\beta$ from the uniform sampling of $[41, 51]$ and $[11, 21]$.
Note that the testing sets used in all evaluations are simulated with the same parameter settings for a fair comparison.
\begin{table}[h]
       
        \small
        %\scriptsize
        \centering
        \caption{Image quality results on the  \textbf{VIS-TH} dataset.}
         \vspace{-1mm}
        \tabcolsep=0.1cm
        \resizebox{\columnwidth}{!}{
                %\hspace{-0.5cm}
                \begin{tabular}{l|cc|c|cc}
                        \hline
                        Methods       & LPIPS$\downarrow$   &NIQE$\downarrow$    & Deg.$\uparrow$   & PSNR$\uparrow$  & SSIM$\uparrow$  \\ \hline \hline
                        TH+TB         &0.7162  & 16.547 &32.21 &6.59 &0.3842  \\ \hline
                        One-Stage~\cite{yang2020hifacegan}       & {0.3355} & 6.532& {50.76} &\textbf{17.64}  & \textbf{0.7208} \\
                        Two-Stage          & 0.3740 &{5.967} &50.04 &15.92  & {0.6941} \\ 
                        TH Only            &  0.4243 & 6.445& 40.78&15.59  &0.6819  \\ \hline
                        \textbf{Ours} &\textbf{0.3127}  &\textbf{5.547} &\textbf{51.68} &{16.91} &0.6836 \\ \hline
                        
        \end{tabular}}\label{tab:q1}
        \vspace{-1mm}
\end{table}

\begin{figure*}[htbp]
  \begin{subfigure}[t]{0.16\linewidth}
    \captionsetup{justification=centering, labelformat=empty, font=scriptsize}
    \includegraphics[width=1\linewidth]{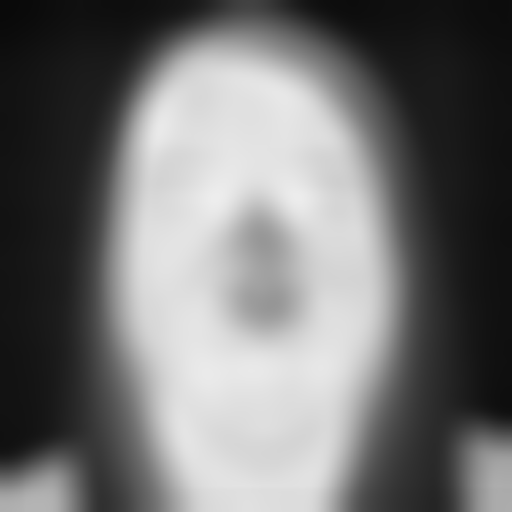}
    \includegraphics[width=1\linewidth]{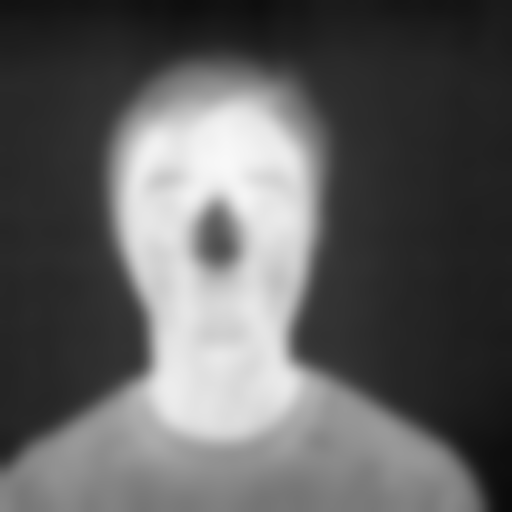}
    \caption{Turbulence-degraded Thermal}
  \end{subfigure}
  \begin{subfigure}[t]{.16\linewidth}
     \captionsetup{justification=centering, labelformat=empty, font=scriptsize}
    \includegraphics[width=1\linewidth]{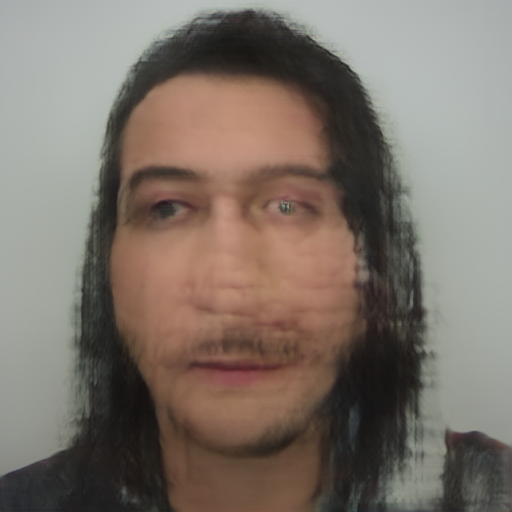}
    \includegraphics[width=1\linewidth]{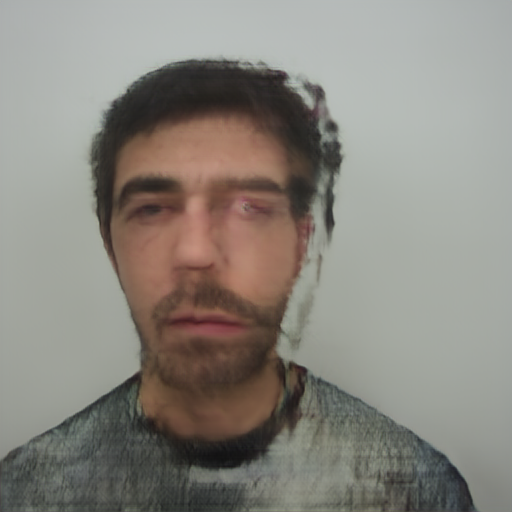}
    \caption{One-Stage}
  \end{subfigure}
  \begin{subfigure}[t]{.16\linewidth}
    \includegraphics[width=1\linewidth]{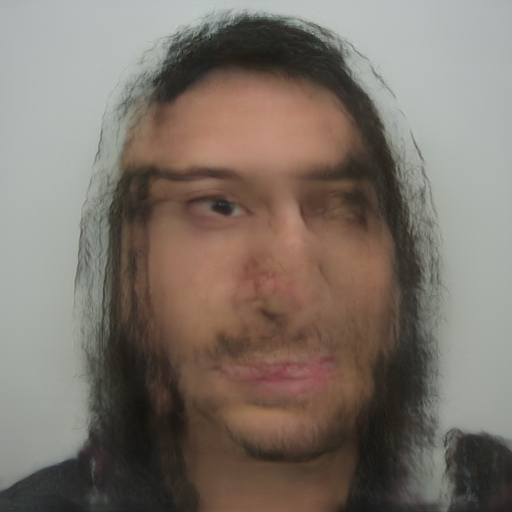}
    \includegraphics[width=1\linewidth]{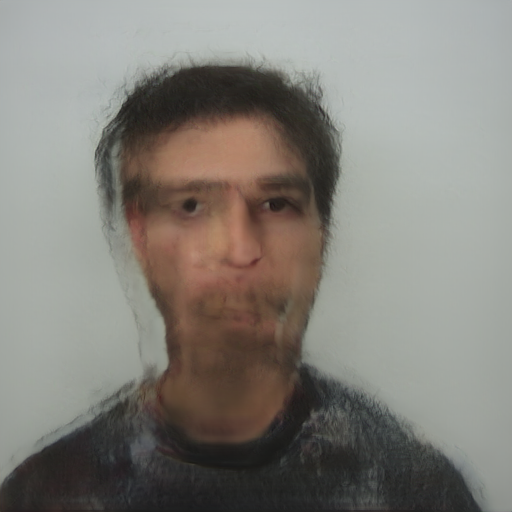}
    \caption{Two-Stage}
  \end{subfigure}
  \begin{subfigure}[t]{.16\linewidth}
    \captionsetup{justification=centering, labelformat=empty, font=scriptsize}
    \includegraphics[width=1\linewidth]{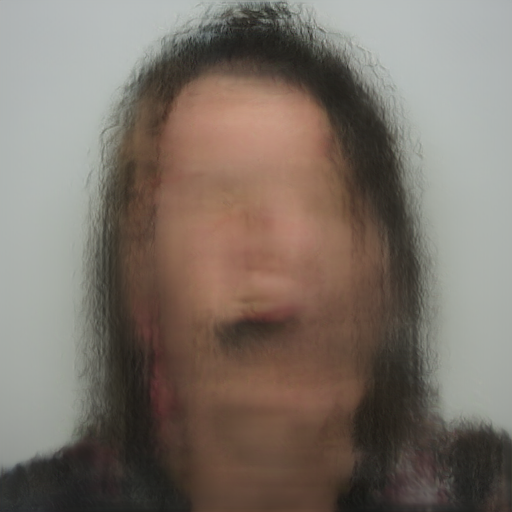}
    \includegraphics[width=1\linewidth]{results/th_1.png}
    \caption{Thermal Only}
  \end{subfigure}
  \begin{subfigure}[t]{.16\linewidth}
    \includegraphics[width=1\linewidth]{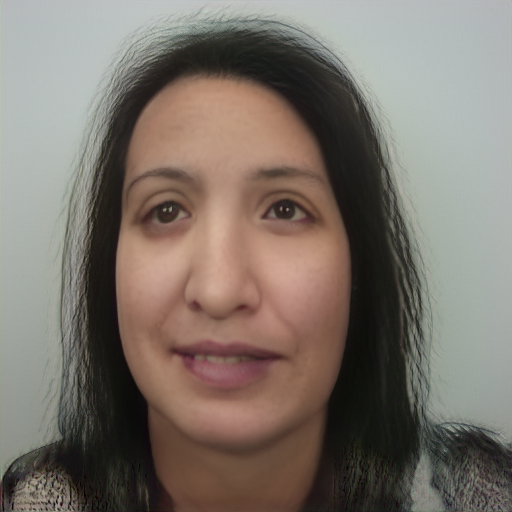}
    \includegraphics[width=1\linewidth]{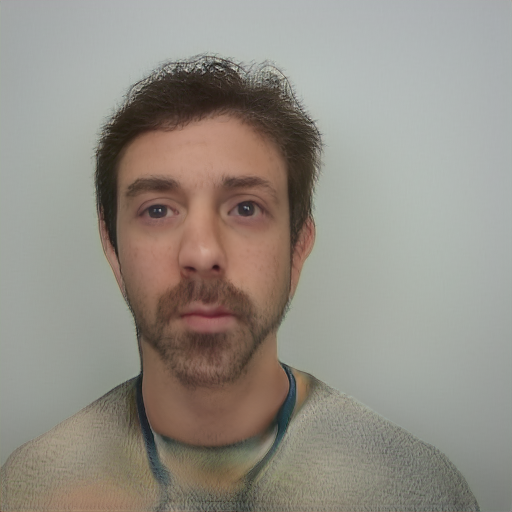}
    \caption{Ours}
  \end{subfigure}
  \begin{subfigure}[t]{.16\linewidth}
    \includegraphics[width=1\linewidth]{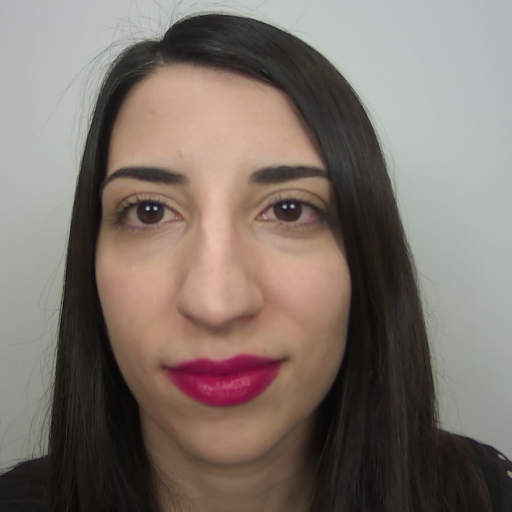}
    \includegraphics[width=1\linewidth]{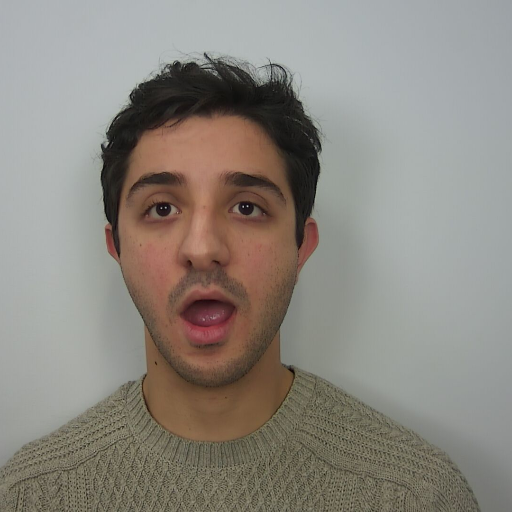}
    \caption{Reference}
  \end{subfigure}
  \hfill
 \vskip-8pt
 \caption{Visualization results of compared methods on the thermal images with simulated turbulence effects.}
 \vskip-10pt
  \label{fig:realcomp}
\end{figure*}

\begin{table}[htbp] 
    \begin{center}
    \centering
    \vskip-5pt
    \caption{Verification results on the \textbf{VIS-TH} dataset.}
   \vspace{-1mm}
   \tabcolsep=0.1cm
    \resizebox{\columnwidth}{!}{
    \begin{tabular}{lccc}
        \hline
         Method & Rank-1 & VR@FAR=1\% & VR@FAR=0.1\% \\
         %Method & Rank-1 & VR1\% & VR0.1\% \\
        \hline
        LightCNN~\cite{wu2018light}& 18.10 & 0.48 & 0 \\ \hline
        One-Stage~\cite{yang2020hifacegan} &41.43  & 7.14& 2.86\\
        Two-Stage &32.38 &4.76 &0 \\
        Direct &{12.38} &{0} &0 \\ \hline
        \textbf{Ours}& \textbf{48.10} &\textbf{10.95} &\textbf{3.33} \\
        \hline
    \end{tabular}}
    \label{tab:r1}
    \end{center}
   \end{table}
\begin{table}[htbp]
        \small
        %\scriptsize
        \centering
        \caption{Image quality results on the \textbf{ARL-VTF} dataset.}
         \vspace{-1mm}
        \tabcolsep=0.1cm
        \resizebox{\columnwidth}{!}{
                %\hspace{-0.5cm}
                \begin{tabular}{l|cc|c|cc}
                        \hline
                        Methods       & LPIPS$\downarrow$   &NIQE$\downarrow$    & Deg.$\uparrow$   & PSNR$\uparrow$  & SSIM$\uparrow$  \\ \hline \hline
                        TH+TB         &0.7111  &17.52 &26.98 &5.85 & 0.3365 \\ \hline
                        One-Stage~\cite{yang2020hifacegan}       &0.3963&10.58  &50.71 &17.88 & 0.7711 \\
                        Two-Stage         &0.2819 &8.564 &56.35 &18.31  &0.7848  \\ 
                        Direct            &0.3786  &8.773 &42.86 &17.56  &0.7615  \\ \hline
                        \textbf{Ours} & \textbf{0.2185}  & \textbf{6.093}& \textbf{61.99}&\textbf{19.06}  &\textbf{0.7586} \\ \hline
                        
        \end{tabular}}
        \label{tab:q2}
        \vspace{-5mm}
\end{table}

\begin{table}[t] 
    \begin{center}
    \centering
    \caption{Verification results on the \textbf{ARL-VTF} dataset.}
   \vspace{-1mm}
   \tabcolsep=0.1cm
    \resizebox{\columnwidth}{!}{
    \begin{tabular}{lccc}
        \hline
         Method & Rank-1 & VR@FAR=1\% & VR@FAR=0.1\% \\
         %Method & Rank-1 & VR1\% & VR0.1\% \\
        \hline
        LightCNN~\cite{wu2018light}&5.69  &5.38 &0.05 \\ \hline
        One-Stage~\cite{yang2020hifacegan} &19.90  &11.47 &4.47 \\
        Two-Stage &29.54 &22.34 &10.46 \\
        Direct &15.84  &7.51 &0.29 \\ \hline
        \textbf{Ours}&\textbf{46.40} &\textbf{25.58} &\textbf{10.96} \\
        \hline
    \end{tabular}}
    \label{tab:r2}
    \end{center}
    \vspace{-5mm}
\end{table}
\subsection{Results on VIS-TH Dataset}
We evaluate our method (denoted as LRTT) on the VIS-TH dataset and compare it with previous state-of-the art face hallucination approach HifaceGAN~\cite{yang2020hifacegan} trained on the same simulated dataset (denote as \textit{One-Stage}). We further introduce a \textit{Two-Stage} strategy, which first reconstructs a clear thermal image via the state-of-the-art turbulence removal approach TDRN~\cite{yasarla_learning_2021} and then translates the thermal image to visible domain using HifaceGAN. To showcase the difficulty induced by the turbulence, we also report results of directly translating the degraded thermal image to visible domain using HifaceGAN (denote as \textit{Direct}).  

\noindent\textbf{Image Quality Results.} Visual results are shown in Figure 3. From this figure, one can see that our approach can synthesize the most clear and accurate faces. In contrast, \textit{Direct} method reconstructs faces with severe artifacts and distortions. Training with the simulation apparently improves the synthesis quality, but the results from \textit{One-Stage} are still very blurry. Moreover, the \textit{Two-Stage} baseline failed to generate high quality faces and yields the worse results compared to \textit{One Stage}, due to the error accumulation as discussed in Section 1. We report the quantitative results in Table \ref{tab:q1}. Our approach achieves the highest performances in all metrics. It is worth noting that our method also achieves the best Deg. score, which indicates our approach can better preserve the identity information, which is crucial for accurate face verification. 

\noindent \textbf{Face Verification Results.} In Table \ref{tab:r1}, we report face verification results. When comparing with HifaceGAN and \textit{Two-Stage}, our method achieves the best performance under all verification metrics. It significantly improves the rank-1 accuracy of the visible domain recognizer LightCNN~\cite{wu2018light} to 48.10\%. This demonstrate its effectiveness in generating high fidelity faces. In contrast, due to the very low synthesis quality, other strategies even reduce the performance of the LightCNN. 

\subsection{Results on the ARL-VTF Dataset}
To further validate the effectiveness of our approach, we conduct experiments on the ARL-VTF dataset. Visual results are shown in Figure 3. ARL-VTF is an easier dataset as it contains more data with variations only in expressions. Therefore, baseline methods can generate faces with reasonable quality. However, although they can produce a rough outline and major facial components, they failed to recover the detailed facial structures and the output images still contain many noticeable artifacts. Our approach can accurately recover the detailed face structures and achieves the highest synthesis quality. Quantitative results are reported in Table \ref{tab:q2}. Our method performs the best in almost all image quality metrics.  The superiority of our approach in thermal-visible face synthesis can further benefit the face verification accuracy. As shown in Table \ref{tab:r2}, one can see that all methods improve the face recognition accuracy based on LightCNN by a large margin. This is mainly because all baselines can produce the face outline. Since LightCNN is a very powerful visible domain face classifier, this can already improve LightCNN to reach a reasonable performance. However, benefiting from more accurate facial details, our method still yields the best performances in all verification metrics.

\section{Conclusion}
We presented a novel GAN inversion network for end-to-end thermal to visible image translation, where the input suffers from atmospheric turbulence.
Compared with the recent thermal image translation approach, two-step turbulence mitigation approach and thermal to visible translation procedure, our method outperforms the other approaches in both visual quality and identity consistency.
Though the evaluation is conducted on the synthetically generated turbulence degraded thermal images, we point out that both the network backbone and data augmentation are thoroughly investigated in real-world cases separately.
Therefore, we believe the proposed method is a new strong baseline for the similar thermal-spectrum translation tasks affected by atmospheric turbulence.

\clearpage
\bibliographystyle{IEEEbib}
{\footnotesize
\bibliography{strings,refs, GANInversion}

\begin{thebibliography}{10}

\bibitem{klare2012heterogeneous}
Brendan~F Klare and Anil~K Jain,
\newblock ``Heterogeneous face recognition using kernel prototype
  similarities,''
\newblock {\em IEEE Transactions on Pattern Analysis and Machine Intelligence},
  vol. 35, no. 6, pp. 1410--1422, 2012.

\bibitem{riggan2016optimal}
Benjamin~S Riggan, Nathaniel~J Short, and Shuowen Hu,
\newblock ``Optimal feature learning and discriminative framework for
  polarimetric thermal to visible face recognition,''
\newblock in {\em IEEE winter conference on applications of computer vision}.
  IEEE, 2016, pp. 1--7.

\bibitem{gong2017heterogeneous}
Dihong Gong, Zhifeng Li, Weilin Huang, Xuelong Li, and Dacheng Tao,
\newblock ``Heterogeneous face recognition: A common encoding feature
  discriminant approach,''
\newblock {\em IEEE Transactions on Image Processing}, vol. 26, no. 5, pp.
  2079--2089, 2017.

\bibitem{choi2012thermal}
Jonghyun Choi, Shuowen Hu, S~Susan Young, and Larry~S Davis,
\newblock ``Thermal to visible face recognition,''
\newblock in {\em Sensing Technologies for Global Health, Military Medicine,
  Disaster Response, and Environmental Monitoring II; and Biometric Technology
  for Human Identification IX}. International Society for Optics and Photonics,
  2012, vol. 8371, p. 83711L.

\bibitem{saxena2016heterogeneous}
Shreyas Saxena and Jakob Verbeek,
\newblock ``Heterogeneous face recognition with cnns,''
\newblock in {\em European Conference on Computer Vision}. Springer, 2016, pp.
  483--491.

\bibitem{fondje2020cross}
Cedric~Nimpa Fondje, Shuowen Hu, Nathaniel~J Short, and Benjamin~S Riggan,
\newblock ``Cross-domain identification for thermal-to-visible face
  recognition,''
\newblock in {\em IEEE International Joint Conference on Biometrics}, 2020, pp.
  1--9.

\bibitem{he2017learning}
Ran He, Xiang Wu, Zhenan Sun, and Tieniu Tan,
\newblock ``Learning invariant deep representation for nir-vis face
  recognition,''
\newblock in {\em Thirty-First AAAI Conference on Artificial Intelligence},
  2017.

\bibitem{he2018wasserstein}
Ran He, Xiang Wu, Zhenan Sun, and Tieniu Tan,
\newblock ``Wasserstein cnn: Learning invariant features for nir-vis face
  recognition,''
\newblock {\em IEEE Transactions on Pattern Analysis and Machine Intelligence},
  vol. 41, no. 7, pp. 1761--1773, 2018.

\bibitem{iranmanesh2018deep}
Seyed~Mehdi Iranmanesh, Ali Dabouei, Hadi Kazemi, and Nasser~M Nasrabadi,
\newblock ``Deep cross polarimetric thermal-to-visible face recognition,''
\newblock in {\em IEEE International Conference on Biometrics}, 2018, pp.
  166--173.

\bibitem{mallat2019cross}
Khawla Mallat, Naser Damer, Fadi Boutros, Arjan Kuijper, and Jean-Luc Dugelay,
\newblock ``Cross-spectrum thermal to visible face recognition based on
  cascaded image synthesis,''
\newblock in {\em IEEE International Conference on Biometrics}, 2019, pp. 1--8.

\bibitem{zhang2019synthesis}
He~Zhang, Benjamin~S Riggan, Shuowen Hu, Nathaniel~J Short, and Vishal~M Patel,
\newblock ``Synthesis of high-quality visible faces from polarimetric thermal
  faces using generative adversarial networks,''
\newblock {\em International Journal of Computer Vision}, vol. 127, no. 6, pp.
  845--862, 2019.

\bibitem{di2018polarimetric}
Xing Di, He~Zhang, and Vishal~M Patel,
\newblock ``Polarimetric thermal to visible face verification via attribute
  preserved synthesis,''
\newblock in {\em IEEE International Conference on Biometrics Theory,
  Applications and Systems}, 2018, pp. 1--10.

\bibitem{di2019polarimetric}
Xing Di, Benjamin~S Riggan, Shuowen Hu, Nathaniel~J Short, and Vishal~M Patel,
\newblock ``Polarimetric thermal to visible face verification via
  self-attention guided synthesis,''
\newblock in {\em IEEE International Conference on Biometrics}, 2019, pp. 1--8.

\bibitem{immidisetti2021simultaneous}
Rakhil Immidisetti, Shuowen Hu, and Vishal~M. Patel,
\newblock ``Simultaneous face hallucination and translation for thermal to
  visible face verification using axial-gan,''
\newblock in {\em IEEE International Joint Conference on Biometrics}, 2021, pp.
  1--8.

\bibitem{riggan2018thermal}
Benjamin~S Riggan, Nathaniel~J Short, and Shuowen Hu,
\newblock ``Thermal to visible synthesis of face images using multiple
  regions,''
\newblock in {\em IEEE Winter Conference on Applications of Computer Vision},
  2018, pp. 30--38.

\bibitem{kolmogorov1991local}
Andrei~Nikolaevich Kolmogorov,
\newblock ``The local structure of turbulence in incompressible viscous fluid
  for very large reynolds numbers,''
\newblock {\em Proceedings of the Royal Society of London. Series A:
  Mathematical and Physical Sciences}, vol. 434, no. 1890, pp. 9--13, 1991.

\bibitem{tatarski2016wave}
Valerian~Ilich Tatarski,
\newblock {\em Wave propagation in a turbulent medium},
\newblock Courier Dover Publications, 2016.

\bibitem{fried1965statistics}
David~L Fried,
\newblock ``Statistics of a geometric representation of wavefront distortion,''
\newblock {\em JoSA}, vol. 55, no. 11, pp. 1427--1435, 1965.

\bibitem{fried1966optical}
David~L Fried,
\newblock ``Optical resolution through a randomly inhomogeneous medium for very
  long and very short exposures,''
\newblock {\em JOSA}, vol. 56, no. 10, pp. 1372--1379, 1966.

\bibitem{fried1978probability}
David~L Fried,
\newblock ``Probability of getting a lucky short-exposure image through
  turbulence,''
\newblock {\em JOSA}, vol. 68, no. 12, pp. 1651--1658, 1978.

\bibitem{yasarla_learning_2020}
Rajeev Yasarla and Vishal~M. Patel,
\newblock ``Learning to {Restore} a {Single} {Face} {Image} {Degraded} by
  {Atmospheric} {Turbulence} using {CNNs},''
\newblock {\em arXiv preprint arXiv:2007.08404}, 2020.

\bibitem{lau_atfacegan_2021}
Chun~Pong Lau, Carlos~D. Castillo, and Rama Chellappa,
\newblock ``{ATFaceGAN}: {Single} {Face} {Semantic} {Aware} {Image}
  {Restoration} and {Recognition} {From} {Atmospheric} {Turbulence},''
\newblock {\em IEEE Transactions on Biometrics, Behavior, and Identity
  Science}, vol. 3, no. 2, pp. 240--251, 2021.

\bibitem{yasarla_learning_2021}
Rajeev Yasarla and Vishal~M. Patel,
\newblock ``Learning to {Restore} {Images} {Degraded} by {Atmospheric}
  {Turbulence} {Using} {Uncertainty},''
\newblock in {\em {IEEE} {International} {Conference} on {Image} {Processing}},
  2021, pp. 1694--1698.

\bibitem{nair_confidence_2021}
Nithin~Gopalakrishnan Nair and Vishal~M. Patel,
\newblock ``Confidence {Guided} {Network} {For} {Atmospheric} {Turbulence}
  {Mitigation},''
\newblock in {\em 2021 {IEEE} {International} {Conference} on {Image}
  {Processing}}, 2021, pp. 1359--1363.

\bibitem{chimitt2020simulating}
Nicholas Chimitt and Stanley~H Chan,
\newblock ``Simulating anisoplanatic turbulence by sampling intermodal and
  spatially correlated zernike coefficients,''
\newblock {\em Optical Engineering}, vol. 59, no. 8, pp. 083101, 2020.

\bibitem{mao2021accelerating}
Zhiyuan Mao, Nicholas Chimitt, and Stanley~H Chan,
\newblock ``Accelerating atmospheric turbulence simulation via learned
  phase-to-space transform,''
\newblock in {\em Proceedings of the IEEE/CVF International Conference on
  Computer Vision}, 2021, pp. 14759--14768.

\bibitem{mei2021ltt}
Kangfu Mei and Vishal~M Patel,
\newblock ``Ltt-gan: Looking through turbulence by inverting gans,''
\newblock {\em arXiv preprint arXiv:2112.02379}, 2021.

\bibitem{karras2020analyzing}
Tero Karras, Samuli Laine, Miika Aittala, Janne Hellsten, Jaakko Lehtinen, and
  Timo Aila,
\newblock ``Analyzing and improving the image quality of stylegan,''
\newblock in {\em IEEE Conference on Computer Vision and Pattern Recognition},
  2020, pp. 8110--8119.

\bibitem{mallat2018benchmark}
Khawla Mallat and Jean-Luc Dugelay,
\newblock ``A benchmark database of visible and thermal paired face images
  across multiple variations,''
\newblock in {\em International Conference of the Biometrics Special Interest
  Group (BIOSIG)}, 2018, pp. 1--5.

\bibitem{poster2021large}
Domenick Poster, Matthew Thielke, Robert Nguyen, Srinivasan Rajaraman, Xing Di,
  Cedric~Nimpa Fondje, Vishal~M Patel, Nathaniel~J Short, Benjamin~S Riggan,
  Nasser~M Nasrabadi, et~al.,
\newblock ``A large-scale, time-synchronized visible and thermal face
  dataset,''
\newblock in {\em IEEE Winter Conference on Applications of Computer Vision},
  2021, pp. 1559--1568.

\bibitem{duan2020cross}
Boyan Duan, Chaoyou Fu, Yi~Li, Xingguang Song, and Ran He,
\newblock ``Cross-spectral face hallucination via disentangling independent
  factors,''
\newblock in {\em IEEE Conference on Computer Vision and Pattern Recognition},
  2020, pp. 7930--7938.

\bibitem{zhang2018unreasonable}
Richard Zhang, Phillip Isola, Alexei~A Efros, Eli Shechtman, and Oliver Wang,
\newblock ``The unreasonable effectiveness of deep features as a perceptual
  metric,''
\newblock in {\em IEEE Conference on Computer Vision and Pattern Recognition},
  2018, pp. 586--595.

\bibitem{mittal2012making}
Anish Mittal, Rajiv Soundararajan, and Alan~C Bovik,
\newblock ``Making a “completely blind” image quality analyzer,''
\newblock {\em IEEE Signal processing letters}, vol. 20, no. 3, pp. 209--212,
  2012.

\bibitem{wu2018light}
Xiang Wu, Ran He, Zhenan Sun, and Tieniu Tan,
\newblock ``A light cnn for deep face representation with noisy labels,''
\newblock {\em IEEE Transactions on Information Forensics and Security}, vol.
  13, no. 11, pp. 2884--2896, 2018.

\bibitem{wang2004image}
Zhou Wang, Alan~C Bovik, Hamid~R Sheikh, and Eero~P Simoncelli,
\newblock ``Image quality assessment: from error visibility to structural
  similarity,''
\newblock {\em IEEE Transactions on Image Processing}, vol. 13, no. 4, pp.
  600--612, 2004.

\bibitem{kingma2015adam}
Diederik~P Kingma and Jimmy Ba,
\newblock ``Adam: A method for stochastic optimization,''
\newblock in {\em International Conference on Learning Representations}, 2015.

\bibitem{yang2020hifacegan}
Lingbo Yang, Shanshe Wang, Siwei Ma, Wen Gao, Chang Liu, Pan Wang, and Peiran
  Ren,
\newblock ``Hifacegan: Face renovation via collaborative suppression and
  replenishment,''
\newblock in {\em ACM International Conference on Multimedia}, 2020, pp.
  1551--1560.

\end{thebibliography}
}
\end{document}